\documentclass[]{opt2024} 

\title[Dual Space Training for GANs]{Dual Space Training for GANs: A Pathway to Efficient and Creative Generative Models}

\optauthor{%
\Name{Beka Modrekiladze} \Email{bekam@cmu.edu}\\
\addr Carnegie Mellon University, Pittsburgh, PA 15213}

\begin{document}

\maketitle

\begin{abstract}%
Generative Adversarial Networks (GANs) have demonstrated remarkable advancements in generative modeling; however, their training is often resource-intensive, requiring extensive computational time and hundreds of thousands of epochs. This paper proposes a novel optimization approach that transforms the training process by operating within a dual space of the initial data using invertible mappings, specifically autoencoders. By training GANs on the encoded representations in the dual space, which encapsulate the most salient features of the data, the generative process becomes significantly more efficient and potentially reveals underlying patterns beyond human recognition. This approach not only enhances training speed and resource usage but also explores the philosophical question of whether models can generate insights that transcend the human intelligence while being limited by the human-generated data.
\end{abstract}

\section{Introduction}

\subsection{Context and Motivation}
Generative Adversarial Networks (GANs) have become a cornerstone in generative modeling due to their ability to learn complex distributions and generate realistic data \cite{goodfellow2014generative}. However, training GANs remains a computationally expensive and time-consuming task, especially with the high demands of modern applications. Current approaches often involve direct learning from datasets, requiring GANs to infer intricate data distributions that are inherently complex.

\subsection{Problem Statement}
\begin{figure}[h]
\centering
\includegraphics[width=0.8\textwidth]{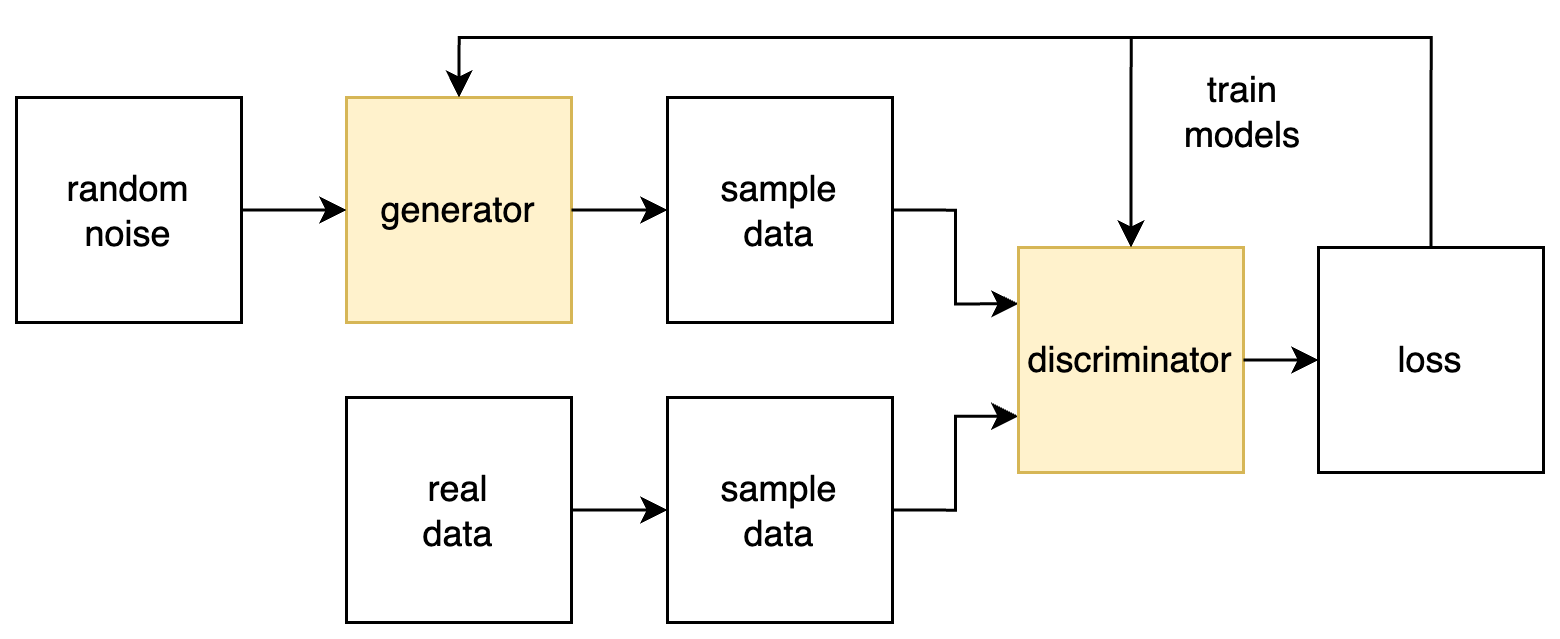}
\caption{Generative Adversarial Network}
\label{GAN}
\end{figure}

The traditional training paradigm for GANs (Figure \ref{GAN}) involves learning directly from data distributions, which is inefficient and demands high computational resources. This inefficiency becomes a bottleneck in large-scale and real-time applications.

\subsection{Proposed Solution}
\begin{figure}[h]
\centering
\includegraphics[width=0.6\textwidth]{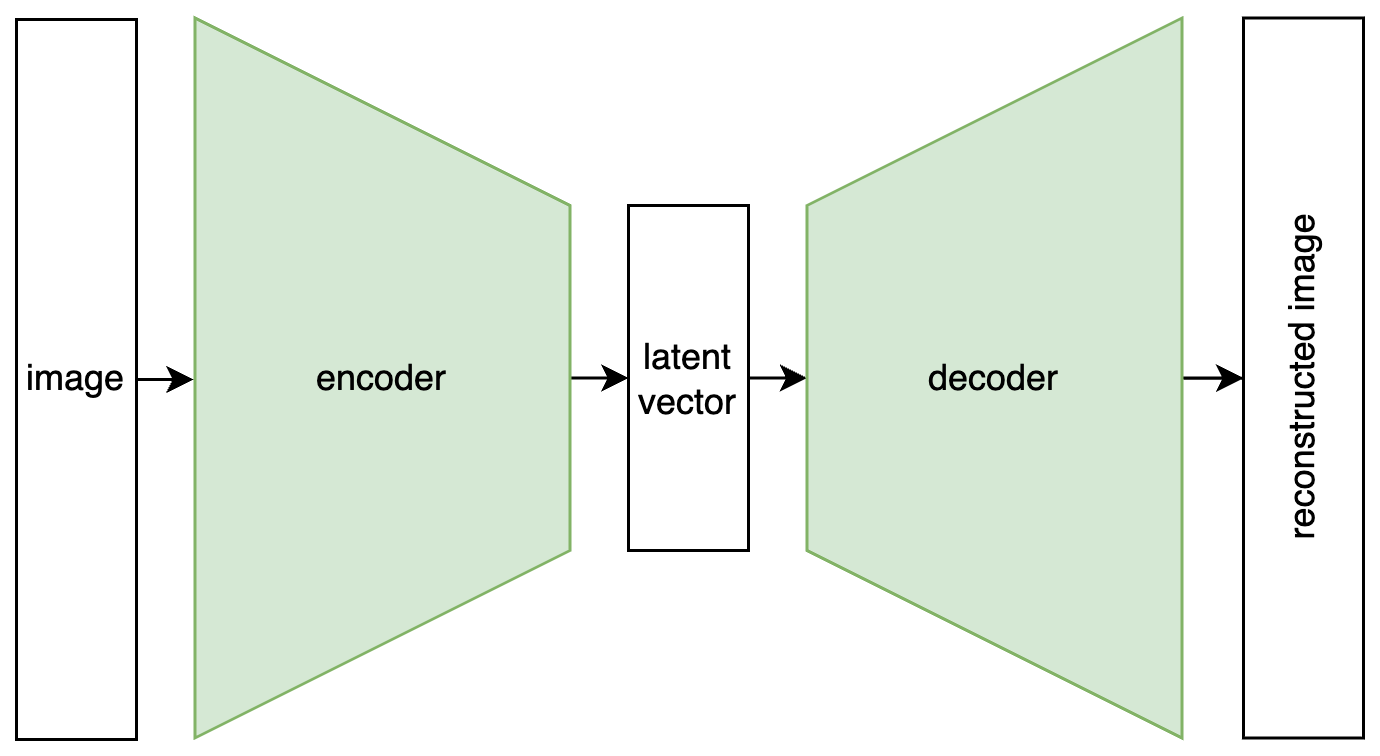}
\caption{Autoencoder}
\label{auto}
\end{figure}

We propose shifting the training of GANs to a dual space of the initial data using invertible mappings, such as those provided by autoencoders \cite{kingma2013auto} (Figure \ref{auto}). By training GANs on encoded representations in a more structured and compressed form, we can drastically reduce the training complexity and resource requirements. It is important to note that this approach is not limited to autoencoders; various dual spaces can be constructed using different invertible mappings, depending on the properties desired for specific applications.

\subsection{Philosophical Insight}
With AI providing more flexible neural networks and having access to different time flows that do not require millions of years of evolution, there is hope to achieve intelligence beyond human capabilities and discover "outer" logics—logics that go beyond what the human brain, constrained by its neural architecture, can generate or even comprehend. While the implications of such possibilities are astonishing, they are met with fair skepticism. The main argument is that, although AI can have arbitrary architectures, it is trained on human-generated data, so it might not venture far from human intelligence. However, in our proposed scenario, generative models will train on dual spaces of the data, which, while observed in the human realm, are fed in a form that encapsulates their fundamental properties. This opens up the potential for uncovering insights that extend beyond the direct confines of human comprehension.

\section{Methodology}
\subsection{Dual Space Concept}
We introduce the concept of dual spaces, where data is transformed into a representation that retains the most salient features while discarding redundant or non-essential information. An invertible map is created between the original and the dual space, allowing for efficient data manipulation and training.

\subsection{Application of Autoencoders}
Autoencoders are used as a practical mapping tool, compressing data into a hidden layer with a much lower dimensionality \cite{kingma2013auto}. The encoder and decoder act as the map and inverse map, respectively, allowing GANs to operate on compressed data and later translate generated outputs back into the original data format.

\subsection{Training in Dual Space}
\begin{figure}[h]
\centering
\includegraphics[width=1\textwidth]{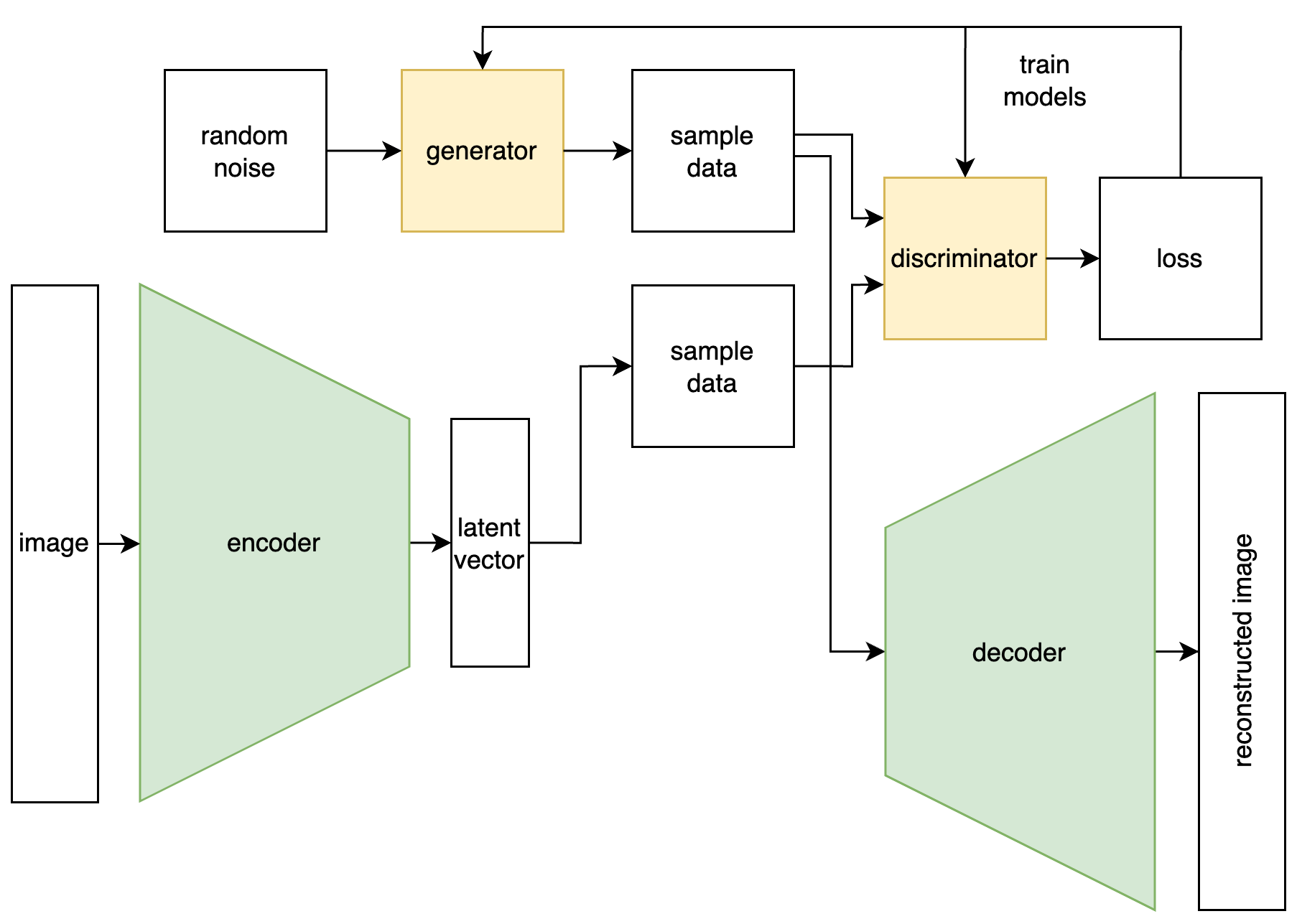}
\caption{Proposed Pipeline}
\label{autogan}
\end{figure}

GANs are trained on the encoded data, which is orders of magnitude smaller and simpler than the original data \cite{radford2015unsupervised}. After the GAN generates new samples in the dual space, the decoder reconstructs these into data-like points in the original space (Figure \ref{autogan}).

\subsection{Advantages}
This approach can significantly reduce training times and computational load. By operating in a space that captures essential features, GANs might identify and generate patterns that are not explicitly present in the raw data, potentially leading to new discoveries.

\section{Experimental Results}
To evaluate the effectiveness of our proposed approach, we conducted experiments using a toy model aimed at generating images of dogs. We compared two training methods: our approach, where an autoencoder was first trained on dog images and then a GAN was trained on the latent features, and the conventional method where the GAN was directly trained on the image dataset.

Our approach demonstrated a significant reduction in training time, taking approximately 100 times less time compared to the standard GAN training method. Beyond the efficiency gains, our method exhibited a deeper understanding of the underlying "dog space." To test this, we deliberately excluded certain dog breeds from the training data. The standard GAN, trained directly on the images, was only able to recreate the breeds it had seen during training. In contrast, the GAN trained on the latent features generated by our approach was capable of producing images of dogs that resembled the missing breeds, despite not having directly seen them in the training data. This suggests that by learning from a more fundamental representation, our approach enables the GAN to extrapolate and generate outputs that go beyond the limitations of the initial dataset.

\section{Discussion and Conclusion}

\subsection{Future Work}
The proposed approach of training GANs in dual spaces opens several avenues for further research and development. Future work will involve exploring alternative invertible mappings beyond autoencoders to identify those that best capture the essential features of specific datasets, optimizing both training efficiency and the quality of generated outputs. Additionally, investigating the impact of different dual space constructions on various types of data, such as high-dimensional, sparse, or noisy datasets, could provide deeper insights into the versatility and limitations of this method.

Further studies could also focus on the theoretical foundations of dual space training, aiming to formally characterize the properties that make certain dual spaces more conducive to efficient learning. This would include analyzing how different mappings affect the convergence behavior of GANs and the diversity of generated samples.

Finally, expanding this framework to other generative models and evaluating its effectiveness across a broader range of applications, including unsupervised learning, anomaly detection, and reinforcement learning, would provide a comprehensive understanding of its potential. Such explorations will help determine the generalizability of the dual space concept and its capacity to push the boundaries of current generative modeling paradigms.

\subsection{Discussion}
Our results strongly indicate that this approach has the potential not only to optimize and accelerate generative models but also to open new doors for creating outputs that extend beyond the boundaries of human intelligence. The dual space training framework allows GANs to work with more fundamental representations of data, which encapsulate essential features rather than the raw, human-generated data itself. This raises the possibility of models uncovering underlying principles or generating novel insights that transcend direct human input.

For instance, if a model were trained solely on data related to biology, it might eventually uncover insights akin to rediscovering chemistry, and with chemistry, it might hint at principles from quantum physics. Similarly, when GANs generate images of faces, they are not merely replicating seen examples but are tapping into the underlying distribution of the "face space." This distribution could be interpreted as a hidden theory of biology, akin to extracting DNA-like representations that define the structure and function of faces. By manipulating these encoded structures, GANs might generate new faces that reflect variations not explicitly seen in the training data.

This suggests a pathway for generative models to explore creative outputs that are not directly constrained by the limitations of human-generated data. In essence, this approach may represent a unique step towards achieving creativity and intelligence that goes beyond human capacity, leveraging the power of abstract, encoded spaces to reveal insights that are yet to be discovered by traditional means.

\subsection{Conclusion}
The implications of our approach extend beyond optimizing GAN training; they challenge the conventional boundaries of what generative models can achieve when detached from the constraints of direct human data. By leveraging dual spaces, our framework not only accelerates the learning process but also suggests a novel route toward achieving creative outputs that could surpass human comprehension. This work paves the way for further exploration into how AI, when guided by more abstract and fundamental representations, can contribute to new scientific discoveries and redefine our understanding of creativity in machine learning.

\bibliography{sample}


\end{document}